\documentclass[runningheads]{llncs}

\usepackage{graphicx}
\usepackage{amsmath}
\usepackage{bm}
\usepackage{url}
\usepackage{stmaryrd}
\usepackage{balance}
\usepackage{amssymb}
\usepackage{pgfplots}
\usepackage{pifont}
\usepackage{float}
\usepackage[misc]{ifsym}
\usepackage{hyperref}

\usepackage{etoolbox} 
\makeatletter
\patchcmd{\ps@headings}{\rlap{\thepage}}{}{}{}
\patchcmd{\ps@headings}{\llap{\thepage}}{}{}{}
\makeatother
\usepackage{epsfig}
\usepackage{booktabs}
\usepackage{pgffor}
\usepackage{tikz}
\usepackage{multirow}
\usepackage{mathrsfs}
\usepackage{bbm}
\usepackage{helvet}
\usepackage{courier}
\usepackage{threeparttable}
\usepackage{algpseudocode} 
\usepackage{arydshln}
\usepackage{mathrsfs}
\usepackage{rotating} 
\usepackage{adjustbox}
\usepackage{makecell}
\usepackage{diagbox} 
\usepackage[skip=0pt]{subcaption}
\usepackage[skip=0pt]{caption}
\usepackage{subfloat}

\usepackage{textcomp}
\usepackage{xcolor}
\usepackage{bigstrut}

\usepackage{pifont}
\usepackage{colortbl}

\usepackage{enumitem}
\usepackage{color}
\usepackage{tcolorbox}
\tcbuselibrary{breakable}
\tcbuselibrary{skins}
\newcommand{\spara}[1]{\vspace{2mm}\noindent\textbf{#1.}}

\newcommand{\repeatthanks}{\textsuperscript{\thefootnote}}

\title{Large Language Models as Topological Structure Enhancers for Text-Attributed Graphs}

\author{Shengyin Sun\inst{1}\thanks{Two authors contributed equally to this work} \and
        Yuxiang Ren\inst{2}\repeatthanks \and
	  Jiehao Chen\inst{3} \and
	  Chen Ma\inst{1}\textsuperscript{(\Letter)}}
\authorrunning{S. Sun et al.}
\institute{\hspace{-1em}
Department of Computer Science, City University of Hong Kong, Hong Kong, China\\
	\email{shengysun4-c@my.cityu.edu.hk, chenma@cityu.edu.hk} \and
	Advance Computing and Storage Lab, Huawei Technologies, Shanghai, China
	\email{renyuxiang1@huawei.com} \and
    China Academy of Industrial Internet, Beijing, China
    \email{chenjiehao@china-aii.com}}

\begin{document}
\titlerunning{Large Language Models as Topological Structure Enhancers for TAGs}
\maketitle

\begin{abstract}
Inspired by the success of Large Language Models (LLMs) in natural language processing (NLP), recent works have begun investigating the potential of applying LLMs in graph learning. However, most existing work focuses on utilizing LLMs as node feature augmenters, leaving employing LLMs to enhance topological structures an understudied problem. In this paper, we are dedicated to leveraging LLMs' text comprehension capability to enhance the topological structure of text-attributed graphs (TAGs). First, we propose using LLMs to help remove/add edges in the TAG. Specifically, we first let the LLM output the semantic similarity between nodes through delicate prompt designs, and then perform edge deletion/addition based on the similarity. Second, we propose using pseudo-labels generated by the LLM to improve graph topology; we introduce pseudo-label propagation as a regularization to guide the graph neural network (GNN) in learning proper edge weights. Finally, we incorporate the two aforementioned graph topological refinements into the GNN training, theoretically justify the benefits of the proposed topology refinements, and perform extensive experiments on real-world datasets to demonstrate the effectiveness of the proposed methods. Code available at \url{https://github.com/sunshy-1/LLM4GraphTopology}.
\end{abstract}

\section{Introduction}
\label{ssy1210:introduction}
Graphs are ubiquitous in various scenarios such as molecular graphs in chemistry~\cite{arxiv:SunYu25} and social networks~\cite{DBLP:conf/ecml/SunMa24}. In the real world, the text-attributed graph (TAG) is also representative, which integrates both graph topological structure and text, enabling them to contain richer semantics. To effectively learn graph-structured data, graph neural networks (GNNs) have emerged. They introduce a message-passing mechanism to aggregate features from neighboring nodes, allowing the simultaneous incorporation of graph topology and node features~\cite{DBLP:conf/nips/HamiltonYing17}.

In the classic GNN pipeline, the text attributes of nodes are usually encoded as shallow embeddings, for example, by Word2Vec~\cite{arxiv:MikolovChen13}. However, shallow embeddings have several limitations, such as the inability to effectively capture sentence-level semantics~\cite{DBLP:conf/ACL/MiaschiOrletta20} and inadequacy in handling the phenomenon of polysemy~\cite{DBLP:journals/scichina/QiuSun20}. To overcome them, recent works have introduced text encoding methods powered by language models (LMs), which obtain contextualized textual embeddings by fine-tuning LMs (e.g., BERT~\cite{DBLP:conf/NAACL/DevlinChang19}) on the specific task.\\
\indent Although LMs have achieved satisfactory results in various tasks, researchers are still dedicated to further increasing the scale of model parameters to capture more complex linguistic patterns~\cite{arxiv:KaplanMcCandlish23,DBLP:conf/prcv/WangWang24,DBLP:conf/MLMI/MaiZhang24}. Recently, this vision has been realized by large language models (LLMs). State-of-the-art LLMs such as ChatGPT~\cite{DBLP:conf/nips/TomB20} and GPT4~\cite{arxiv:OpenAI23}, with hundreds of billions of parameters, have demonstrated remarkable capabilities across a wide range of NLP tasks~\cite{DBLP:conf/ACL/HuChen24,DBLP:conf/coling/XiaoBlanco24}. However, the application of LLM's capabilities to graph-structured data remains relatively underexplored. Early attempts focused on using LLMs to enhance node features and did not consider how to utilize LLMs for perturbing the graph structure. In this paper, we investigate how to harness LLMs to refine the topological structure of TAGs. The key motivation is that \textit{we find TAGs contain many irrational/unreliable edges that can potentially have an adverse effect on the message-passing process in GNN} (we show in Fig. \ref{ssy1101:ratio_different_endpoints_class} that the proportion of potentially unreliable edges in the Cora and Citeseer datasets exceeds 15\%), thus degrading the quality of learned node representations. Specifically, unlike other network architectures such as CNNs and MLPs that individually encode each entity (node), GNNs encode nodes through aggregation/propagation of node features. This means that if two nodes are connected by unreliable edges, their final aggregated representations will inevitably contain some noise. Therefore, it is necessary to remove unreliable edges through graph topology refinement.\\
\indent To mitigate the impact of unreliable edges on model performance, we resort to LLMs for assistance. First, we explore the potential of LLMs in performing edge deletion/addition on TAGs. Specifically, through careful prompt design, we let the LLM output the relatedness between two nodes based on their textual attributes. Based on the obtained semantic-level similarity, we perform edge deletion/addition on TAGs, where we manage to keep (or add) edges between nodes with higher similarity and delete (or not add) edges between nodes with lower similarity. Second, we investigate leveraging pseudo-labels generated by LLMs to improve the graph topology, where we propose to incorporate pseudo-labels generated by LLMs into GNN training. In particular, we introduce the propagation of pseudo-labels as a regularization to guide the model in learning proper edge weights that benefit the separation of different node classes (i.e., enabling nodes with a high probability of belonging to the same class to connect more strongly, while promoting a better separation of nodes with a high probability of belonging to different classes). Finally, we combine the two proposed topology refinement methods with the GNN training. Our contributions are as follows:
\begin{itemize}[leftmargin=*, topsep=2mm]
	\item Integrating LLMs with graph data is challenging, and most of the existing works only focus on using LLMs for feature enhancement. In this paper, we make the first attempt at employing LLMs to improve topological structure.
	\item Two LLM-based graph topology refinement approaches are proposed. One involves using carefully crafted prompt statements to make the LLM generate semantic-level similarity between nodes. The generated similarity can be used to guide the addition of reliable edges and the removal of unreliable ones. The other involves using high-quality pseudo-labels generated by the LLM for pseudo-label propagation. The process of pseudo-label propagation is introduced as a regularization for the GNN training, leading the model to learn proper edge weights, thereby achieving the goal of improving graph topology. Moreover, we theoretically analyze the benefits of the proposed methods.
\end{itemize}

\section{Related Work}
\textbf{Label Propagation}. Traditional LPAs solely rely on labels and do not consider features~\cite{DBLP:conf/nips/ZhouBousquet04}. To better utilize the information contained in node features, various techniques have been proposed for learning edge weights, including using kernel functions~\cite{DBLP:conf/iclr/LiuLee19,DBLP:conf/iclr/ZhuGhahramani03}, minimizing reconstruction error~\cite{DBLP:conf/nips/KarasuyamaKarasuyama13}, and imposing sparsity constraints~\cite{DBLP:conf/iccv/HongLiu09}. Wang et al. proposed an approach using LPA itself as a regularization to help the model learn edge weights~\cite{DBLP:journals/TIOS/WangLeskovec21}, which is different from the adaptive LPA mentioned above using a specific regularization. However, the initial label matrix used for label propagation in  \cite{DBLP:journals/TIOS/WangLeskovec21} is constructed solely using the information from labeled nodes while ignoring the information in unlabeled nodes. In this paper, we utilize LLMs to generate pseudo labels to assist label propagation, which enables the information in unlabeled nodes to be utilized, thereby guiding the model to learn more proper edge weights.\\

\noindent\textbf{Applying LLMs on TAGs}. LLMs~\cite{DBLP:conf/nips/TomB20,arxiv:Touvron23} have demonstrated great potential in various NLP tasks. However, how to apply LLMs to graph-structured data, such as TAGs, remains a challenge~\cite{arxiv:TangYang23,arxiv:ZhuWang24,arxiv:ZhaoLiu23,arxiv:QinWang23}. Chen et al. investigate the potential of LLMs in node classification tasks~\cite{DBLP:conf/kdd/ChenMao24}. 
He et al. propose a novel feature `TAPE' augmented by the LLM~\cite{DBLP:conf/iclr/HeBresson24}, which incorporates the original attributes of nodes in the TAG, predictions made by the LLM on node categories, as well as explanations from the LLM for making such predictions. As shown before, existing methods primarily utilize LLMs to enhance the node features in graphs. However, approaches leveraging LLMs to aid the enhancement of graph topological structures remain relatively underexplored. In this paper, we focus on exploring LLMs' potential in enhancing graph topological structures.

\section{Preliminary}
\label{sun1101:preliminary}
\subsection{Notations}
Sets are denoted by calligraphic letters (e.g., $\mathcal{S}$). $\vert \cdot \vert$ denotes the number of elements in the set (e.g., $\vert\mathcal{S}\vert$). Matrices and vectors are denoted as bold upper case letters (e.g., $\mathbf{X}$) and bold lower case letters (e.g., $\mathbf{x}$), respectively. Superscript $(\cdot)^{\top}$ denotes transpose for matrices or vectors. $\mathbb{R}^{m\times n}$ represents a real matrix space of dimension $m\times n$. $\mathbf{H}[i,:]$ is
the $i$\textit{-th} row of $\mathbf{H}$. $\mathbf{H}{(i,j)}$ represents the $(i,j)$\textit{-th} element in the matrix $\mathbf{H}$. $\|\mathbf{v}\|$ is the norm of $\mathbf{v}$.

\subsection{Text-attributed Graph and Graph Neural Network}
Define the TAG as $\mathcal{G}=\{\mathcal{V}, \mathcal{E}, \mathcal{{X}}, \mathbf{{A}}\}$, where $\mathcal{V}$ is the set of nodes, $\mathcal{E}$ is the set of edges, $\mathcal{{X}}$ is a set of text attributes, and $\mathbf{A}\in \mathbb{R}^{\vert\mathcal{V}\vert\times \vert\mathcal{V}\vert}$ represents an adjacency matrix. More specifically, for each node $v_{i}$ in $\mathcal{V}$, it corresponds to a piece of text $x_{i}\in\mathcal{X}$. If there is an edge between $v_i$ and $v_j$, then $\mathbf{A}_{ij}=1$; otherwise, $\mathbf{A}_{ij}=0$. 
Under the node classification setting, $v_i$ also corresponds to a label $y_i$ that indicates which category the text attribute of node $v_i$ belongs to. Mathematically, the representation of node $u$ at the $l$\textit{-th} GNN layer ${\mathbf{{h}}}_{u}^{l}$ can be expressed as: ${\mathbf{{h}}}_{u}^{l}=f^{l}(\boldsymbol{\Phi};\mathcal{X},\mathbf{A})={\textsc{update}}\left(\mathbf{h}_{u}^{l-1},{\textsc{AGG}}\left(\left\{{\mathbf{{h}}}_{v}^{l-1}, v\in\mathcal{N}_{u}\right\}\right)\right)$, where $\mathcal{N}_{u}$ is the set of neighbors of node $u$, ${\textsc{Agg}}(\cdot)$ is the operation of aggregating messages, $\textsc{update}(\cdot)$ is the operation of updating node $u$'s representation by using its previous representation and aggregated messages. For the sake of simplicity, the ${\textsc{Agg}}(\cdot)$ and $\textsc{update}(\cdot)$ operation in the $l$\textit{-th} layer can be further abstracted into a parameterized function $f^{l}(\cdot)$ with the learnable parameters $\boldsymbol{\Phi}$.

\section{Methodology}
\label{sun1101:Methodology}
\subsection{Necessity of Graph Topology Enhancement}
\noindent\textbf{$\bullet$ The topology enhancement is neglected by existing LLM-GNN works}. A straightforward approach to combining LLMs and GNNs is to utilize the text generation capability of LLMs to enrich the input of GNNs. Employing LLMs as attribute enhancers is exemplified in works such as: (\romannumeral1) TAPE~\cite{DBLP:conf/iclr/HeBresson24}---LLM is used to generate pseudo-labels for each node and is required to provide explanations for generating such pseudo-labels. These pseudo-labels and explanations are concatenated with the original text attributes as input for the GNN; (\romannumeral2) KEA~\cite{DBLP:conf/kdd/ChenMao24}---LLM is employed to extract technical terms from the original text attributes and is tasked with elaborating on these terms in detail. The generated technical terms with descriptions, along with the original text attributes, are then encoded by the GNN. However, these works overlook a crucial point that distinguishes the GNN from other architectures (such as the CNN): GNNs encode nodes through the aggregation/propagation of node features, rather than encoding nodes individually. \emph{In other words, it is challenging to ensure the quality of the final representation (after aggregation) of a node when it is connected to noisy nodes through unreliable edges, no matter how its own features are enhanced.} Therefore, we argue that the LLM-based topology enhancement is as important as the LLM-based attribute enhancement.\\

\begin{figure}[t]
    \centering
    \begin{subfigure}{0.45\textwidth}
        \centering
        \includegraphics[width=\linewidth]{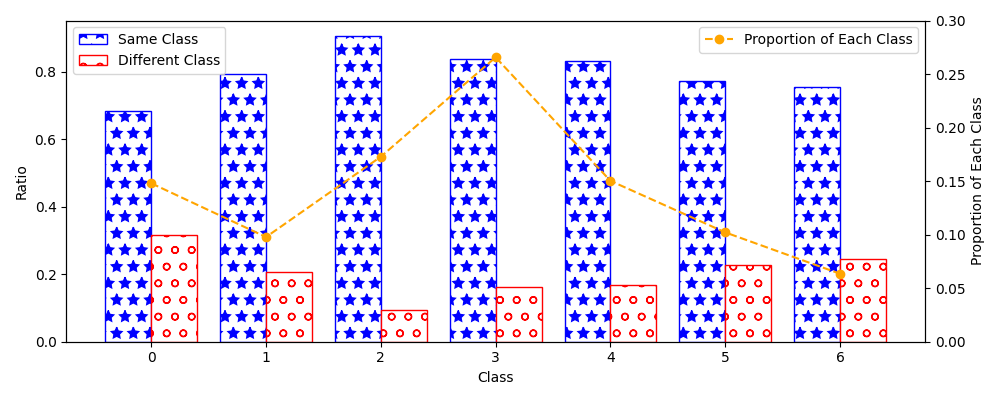}
        \caption{Cora Dataset}
    \end{subfigure}
    \begin{subfigure}{0.45\textwidth}
        \centering
        \includegraphics[width=\linewidth]{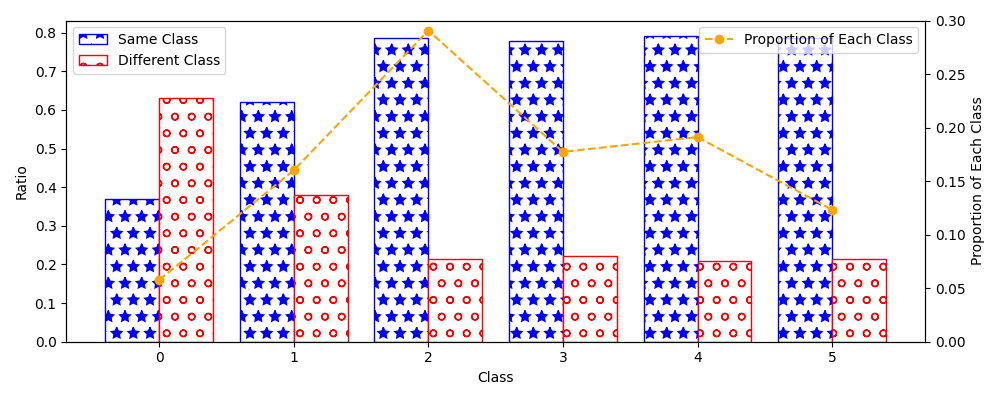}
        \caption{Citeseer Dataset}
    \end{subfigure}
    \caption{The proportion of edge endpoints with the same/different classes.}
    \label{ssy1101:ratio_different_endpoints_class}
\end{figure}

\begin{figure}[t]
    \centering
    \begin{subfigure}{0.35\textwidth}
        \centering
        \includegraphics[width=\linewidth]{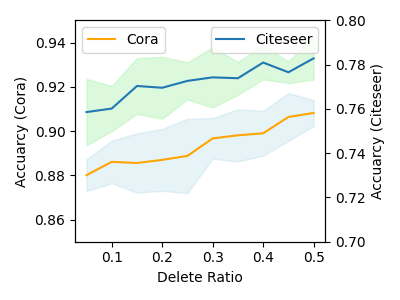}
        \caption{General Setting}
    \end{subfigure}
    \begin{subfigure}{0.35\textwidth}
        \centering
        \includegraphics[width=\linewidth]{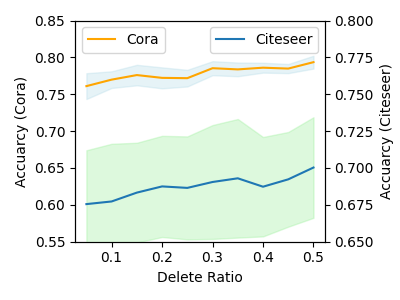}
        \caption{Few-shot Setting}
    \end{subfigure}
    \caption{The impact of removing potentially unreliable edges on accuracy.}
    \label{ssy1101:removing_the_unreliable_egdes}
\end{figure}

\noindent\textbf{$\bullet$ Observations from public datasets}. We investigate the proportion of potentially unreliable edges\footnote{A potentially unreliable edge refers to an edge connecting two endpoints belonging to distinct classes (assuming that we know the true labels of all nodes).} in public datasets, as shown in Fig.~\ref{ssy1101:ratio_different_endpoints_class}. Overall, the proportion of potentially unreliable edges cannot be ignored (averaging over 15\%). Taking Citeseer as an example, among edges where at least one endpoint belongs to the class ``0'', the proportion of potentially unreliable edges exceeds 60\%. This suggests that nodes of class ``0'' exhibit a high probability of connecting to unrelated nodes, which, in turn, can lead to blurred boundaries between the representations of nodes with class ``0'' and those of nodes with other classes after feature aggregation. To further demonstrate the potential of topology enhancement, we explore the impact of removing these potentially unreliable edges. As shown in Fig. \ref{ssy1101:removing_the_unreliable_egdes}, the accuracy steadily improves with the increasing delete ratio, which aligns with our previous analysis.


\begin{figure*}[t]
    \centering
    \includegraphics[width=4.69in]{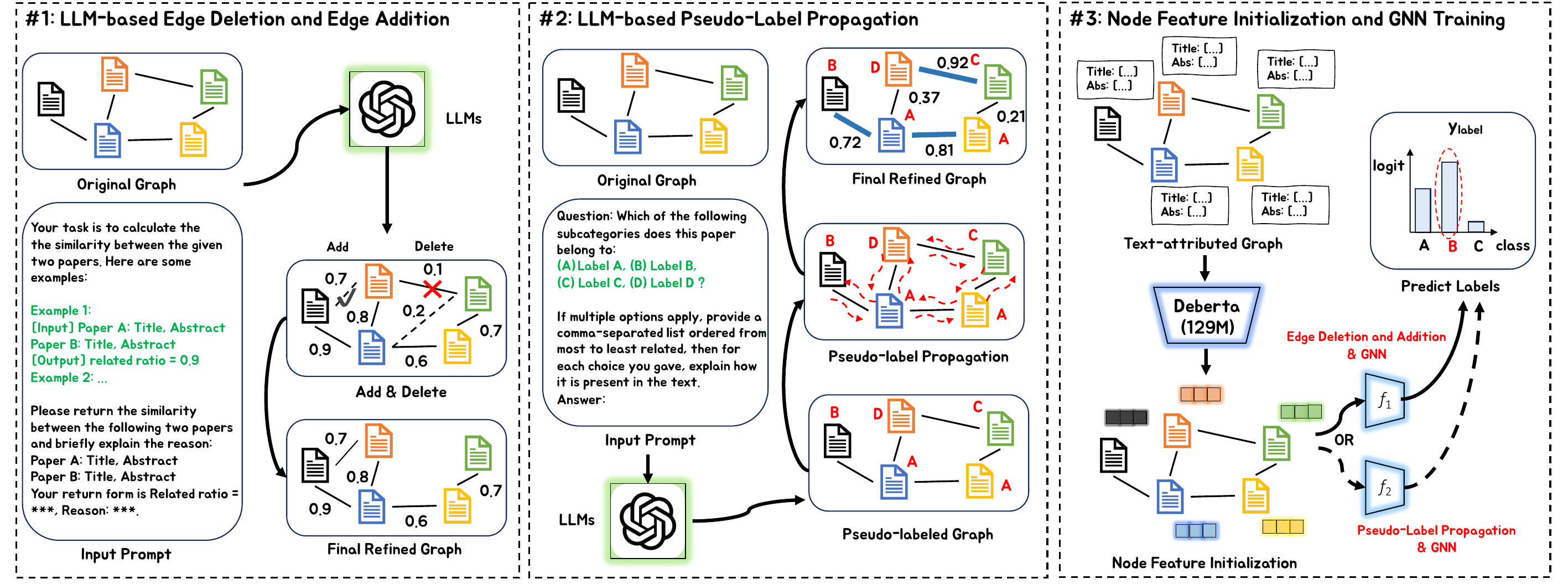}
    \caption{The overview of the proposed methods. (\#1) LLM-based edge deletion/addition. (\#2) LLM-based pseudo-label propagation. (\#3) Integration into GNN training.}
    \label{ssy1101:overall_framework}
\end{figure*}

\subsection{LLM-based Edge Deletion and Addition}
\label{ssy1101:LLM_A_D}
LLMs have introduced a novel paradigm called ``\emph{pre-train, prompt, and predict}''. Under this paradigm, the LLM first undergoes pre-training in large-scale text datasets to learn general knowledge. Then, instead of fine-tuning the pre-trained model for each downstream task, the pre-trained model is prompted with a human-readable prompt according to the specific downstream task. The model generates the output based on the prompt and given input, without any task-specific fine-tuning of the model parameters.

From the perspective of the LLM's input/output, the edge deletion/addition shares the same form of prompt paradigm. Specifically, we need to input the texts of the two endpoints (corresponding to a certain edge) into the LLM and guide the LLM to output its judgment on deleting or adding the edge, i.e.,
\begin{equation}
    \label{sun1101:LLM_prompt}
    r_{ij} = {\rm{LLM}}({\rm{Prompt}}(x_i,x_j)),\quad\quad x_i,x_j\in\mathcal{X},
\end{equation}
where $x_i$ and $x_j$ are text attributes corresponding to the endpoints, ${\rm{Prompt}}(\cdot)$ is the designed prompt format, and $r_g$ is the response text generated by the LLM.  The well-designed prompt structure ${\rm{Prompt}}(\cdot)$ for the edge deletion/addition is:

\begin{tcolorbox}[breakable,enhanced jigsaw,arc=0pt, leftrule=0pt, rightrule=0pt,colback=gray!5]
Your task is to calculate the similarity between the given two papers. Here are some examples where the input is the information for paper A and paper B, and the output is the estimated related ratio between paper A and paper B: \textcolor{purple}{\textbackslash n}\\
\textbf{Example 1:} \textbf{[Input] Paper A}: Title: [paper title ...] {Abstract:} [paper abstract ...] \textbf{Paper B}: Title: [paper title ...] {Abstract:} [paper abstract ...] \textbf{[Output]} Estimated Related Ratio between Paper A and Paper B: 0.x\textcolor{purple}{\textbackslash n}\\
\textbf{Example 2:} ...\textcolor{purple}{\textbackslash n}\\
Please return the similarity between the following two papers and briefly explain the reason: \textbf{Paper A:} Title: [paper title ...] {Abstract:} [paper abstract ...] \textbf{Paper B:} Title: [paper title ...] {Abstract:} [paper abstract ...]. Your return form is: Related ratio = ***, Reason: ***.
\end{tcolorbox}
In the aforementioned prompt structure, we first describe the task to the LLM, that is, to estimate the similarity between two papers based on their titles and abstracts. To help the LLM better understand the task and prevent it from generating meaningless numbers, we then provide multiple examples as references. Finally, we include the text attributes related to the two endpoints (i.e., paper A and paper B), and request the LLM to output its estimations. After obtaining the similarity estimations given by the LLM, we can use a threshold-based method to determine which edges need to be added or removed, i.e.,
\begin{equation}
    \label{ssy1101:AD_adj}
    \mathbf{A}_{\rm A-D}(i,j) = \left\{
\begin{array}{l}
    1, ~~~{\rm{if}~}{{\rm{Extract}}(r_{ij})}>\xi\\
    0, ~~~{\rm{if}~}{{\rm{Extract}}(r_{ij})}\le\xi
\end{array}
\right.
\end{equation}
where $\mathbf{A}_{\rm A-D}$ is the adjacency matrix of the refined graph, ${\rm{Extract}}(\cdot)$ is a function that extracts similarity numbers from LLM's response, $\xi$ is a preset threshold.

\subsection{LLM-based Pseudo-label Propagation}
\label{ssy1101:LLM_LPA}
In this section, we first prompt the LLM to give pseudo-labels based on the text attributes of nodes. For convenience, we take the Citeseer dataset as an example to describe. The prompt structure on the Citeseer dataset is as follows: 
\begin{tcolorbox}[breakable,enhanced jigsaw,arc=0pt, leftrule=0pt, rightrule=0pt,colback=gray!5]

The title and abstract of the paper are as follows:
Title: [paper title ...] 
Abstract: [paper abstract ...]\textcolor{purple}{\textbackslash n}\\
\textbf{Question}: Which of the following subcategories does this paper belong to: (A) Agents, (B) ...? If multiple options apply, provide a comma-separated list ordered from most to least related, then for each choice you gave, explain how it is present in the text.\textcolor{purple}{\textbackslash n}\\
\textbf{Answer}: 
\end{tcolorbox}
After obtaining the response text from the LLM, we choose the category that the LLM thinks the node $v_i$ is most likely to belong to (the category ranked first in the response text) as the pseudo-label $y^p_i$. Inspired by the work~\cite{DBLP:journals/TIOS/WangLeskovec21}, we hope to utilize the pseudo-labels generated by the LLM to optimize edge weights (in a label propagation manner) so that important edges can be highlighted in the message passing of GNNs. Formally, the LLM-based pseudo-label propagation containing $K$ iterations can be written as:
\begin{equation}
    \label{ssy1101:LLM_based_LPA}
    \left\{
    \begin{array}{l}\
    \mathbf{Y}^{(0)}=\left[\mathbf{y}^{(0)}_{1},\dots,\mathbf{y}^{(0)}_{|\mathcal{V}|}\right]^{\top},\\
    \mathbf{Y}^{(k)}=\bar{\mathbf{A}}\mathbf{Y}^{(k-1)}=\left[\mathbf{y}^{(k)}_{1},\dots,\mathbf{y}^{(k)}_{|\mathcal{V}|}\right]^{\top}, ~~~k=1,\cdots,K,
    \end{array}
    \right.
\end{equation}
where $\mathbf{Y}^{(0)}$ is the initial label matrix for the label propagation, $\mathbf{y}^{(0)}_{i}$ for $i=1,\dots,|\mathcal{V}|$ is the one-hot label indicator vector generated according to the $y^p_i$, $\bar{\mathbf{A}}=\mathbf{D}^{-\frac{1}{2}}\mathbf{A}\mathbf{D}^{-\frac{1}{2}}$ is the normalized adjacency matrix (consistent with the normalization method in the GCN~\cite{DBLP:conf/iclr/Thomas16}) with the learnable adjacency matrix $\mathbf{A}$ and the diagonal degree matrix $\mathbf{D}$ for $\mathbf{A}$. The optimal edge weights can be obtained by minimizing the loss of predicted labels by the LLM-based pseudo-label propagation, i.e.,
\begin{equation}
    \label{ssy1101:optimal_edge_weight}
    \mathbf{A}^{\rm{*}} = \arg\min_{\mathbf{A}} \frac{1}{|\mathcal{V}_{\rm train}|}\sum_{v_i \in \mathcal{V}_{\rm train}} \mathcal{L}_{\rm CE}(\mathbf{y}_{i}{^{\text{llm-lpa}}}, \mathbf{y}_{i}),
\end{equation}
where $\mathbf{A}^{*}$ is the optimized graph topology, $\mathcal{L}_{\rm CE}$ is the cross-entropy loss, $\mathcal{V}_{\rm train}$ is the training set containing labeled nodes, $\mathbf{y}_{i}{^{\text{llm-lpa}}}$ is the predicted label distribution of node $v_i$ after $K$ iterations of propagation (i.e, $\mathbf{y}_{i}{^{\text{llm-lpa}}}=\mathbf{y}_{i}^{(K)}$), and $\mathbf{y}_{i}$ is the true one-hot label of node $v_i$. It should be noted that Eq.~\eqref{ssy1101:optimal_edge_weight} merely demonstrates how to utilize the pseudo-labels from LLMs to help optimize the graph topology. In the next section, we will elaborate in more detail on how to integrate the LLM-based pseudo-label propagation into GNN training and train the whole model in an end-to-end manner.

\subsection{LM Fine-tuning and GNN Training}
We fine-tune the pre-trained LM DeBERTa-base~\cite{arxiv:HeLiu20} to encode the text attribute $x_i$ of the node $v_i$ into the vectorial feature, i.e.,
\begin{equation}
\label{ssy1101:LM_encode}
    {\mathbf{h}}_{i}={\rm{LM}}(x_i),~~~v_i\in\mathcal{V}{},x_i\in\mathcal{X},
\end{equation}
where ${\mathbf{h}}_{i}$ is the text embedding for the node $v_i$, $\rm LM(\cdot)$ represents the fine-tuned LM model DeBERTa-base. To extract the most informative textual features tailored for the downstream task, we fine-tune the LM~\cite{DBLP:conf/iclr/HeBresson24}.  More specifically, we apply a multi-layer perceptron (MLP) to the output of the LM and minimize the cross-entropy loss between the LM's predictions and the labels for each node.

For the method of LLM-based edge deletion and edge addition, we can perform the GNN training by directly utilizing the adjacency matrix $\mathbf{A}_{\rm A-D}$ (see Section \ref{ssy1101:LLM_A_D} for details). Formally, the process of message passing and model optimization can be written as follows:
\begin{equation}
    \label{ssy1101:GNN_train_for_A_D}
    \left\{
    \begin{array}{l}\
    \mathbf{H}^{(k)}=\sigma\left(\bar{\mathbf{A}}_{\rm{A-D}}\mathbf{H}^{(k-1)}\mathbf{W}^{(k-1)}\right),~~~k=1,\cdots,K,\\
    \\
    \mathbf{W}^{*}=\underset{\mathbf{W}}{\arg\min} \frac{1}{|\mathcal{V}_{\rm train}|}\underset{v_i \in \mathcal{V}_{\rm train}}{\sum} \mathcal{L}_{\rm CE}(\mathbf{y}_{i}{^{\text{gcn}}}, \mathbf{y}_{i}),
    \end{array}
    \right.
\end{equation}
where $\sigma(\cdot)$ is an activation function, $\mathbf{W}^{(k)}$ is a learnable weight matrix in the k\textit{-th} layer, $\bar{\mathbf{A}}_{\rm{A-D}}$ is the normalized adjacency matrix for ${\mathbf{A}}_{\rm{A-D}}$, $\mathbf{H}^{(k)}$ is the output matrix of the k\textit{-th} layer (the initialization of $\mathbf{H}$ is based on the Eq.~\eqref{ssy1101:LM_encode}, i.e., $\mathbf{H}^{(0)}=\left[\mathbf{h}_1,\cdots,\mathbf{h}_{|\mathcal{V}|}\right]^{\top}$), $\mathbf{\hat y}_{i}{^{\text{gcn}}}$ represents the predicted label
distribution of node $v_i$ by using GCN in Eq.~\eqref{ssy1101:GNN_train_for_A_D}, $\mathbf{W}$ denotes learnable parameters in the whole model, and $\mathbf{W}^{*}$ is the optimized parameters.

For the LLM-based pseudo-label propagation, the process of model optimization and topology optimization can be written as:
\begin{equation}
    \label{ssy1101:llm_lpa_loss}
    \begin{split}
    \begin{aligned}
    \mathbf{W}^{*},\mathbf{A}^{*} = &\underset{\mathbf{W},\mathbf{A}}{\arg\min} \frac{1}{|\mathcal{V}_{\rm train}|}\underset{v_i \in \mathcal{V}_{\rm train}}{\sum} \Big[\mathcal{L}_{\rm CE}(\mathbf{y}_{i}{^{\text{gcn}}}, \mathbf{y}_{i})\\
    &+\lambda \mathcal{L}_{\rm CE}(\mathbf{y}_{i}{^{\text{lpa}}}, \mathbf{y}_{i})+\beta \mathcal{L}_{\rm CE}(\mathbf{y}_{i}{^{\text{llm-lpa}}}, \mathbf{y}_{i})\Big],
    \end{aligned}
    \end{split}
\end{equation}
where $\lambda$ and $\beta$ are hyper-parameters, $\mathbf{y}_i^{\rm{lpa}}$ and $\mathbf{y}_i^{\rm{llm-lpa}}$ are two label distributions obtained from two different label propagation methods. Specifically, $\mathbf{y}_i^{\rm llm-lpa}$ is the label distribution obtained from the initial label matrix constructed using the pseudo-labels generated by the LLM (see Eq.~\eqref{ssy1101:LLM_based_LPA} for details), while $\mathbf{y}_i^{\rm lpa}$ is the label distribution obtained from the initial label matrix constructed using the true labels from the training set (with the one-hot encoded labels for nodes outside the training set initialized as default values)~\cite{DBLP:journals/TIOS/WangLeskovec21}. The overall framework is provided in Fig. \ref{ssy1101:overall_framework}.

\subsection{Theoretical Justification}
\label{ssy1101:theory_analysis}
Essentially, the proposed approaches aim to refine the message-passing process by applying hard/soft masks to graphs. In this section, we harness the shrinking theory~\cite{DBLP:conf/LoG/LinKang23,DBLP:journals/TIOS/WangLeskovec21,DBLP:conf/icdm/YangMeng20} to understand the benefits of proposed methods in refining the message passing process. Define the embedding variation \cite{DBLP:conf/DAGM/Zhou05} on graphs as:
\begin{equation}
    \label{ssy1101:variation_measure}
    \mathcal{M}\left(\mathbf{H}^{(k)}\right)=\frac{1}{2}\sum_{i=1}^{\vert\mathcal{V}\vert}\sum_{j=1}^{\vert\mathcal{V}\vert}\left\|\frac{\mathbf{A}(i,j)}{\mathbf{D}(i,i)}\mathbf{h}_{i}^{(k)}-\frac{\mathbf{A}(j,i)}{\mathbf{D}(j,j)}\mathbf{h}_{j}^{(k)}\right\|^2,
\end{equation}
where $\mathbf{A}, \mathbf{D}, \mathbf{H}^{(k)}$ are the previously refined adjacency matrix, diagonal degree matrix, and embedding matrix, respectively. For the convenience, we rewrite $\mathcal{M}\left(\mathbf{H}^{(k)}\right)$ as $\sum_{i=1}^{\vert\mathcal{V}\vert}\mathcal{M}_{i}\left(\mathbf{h}^{(k)}_{i}\right)$. Utilizing the Taylor's theorem, we have
\begin{equation}
    \label{ssy1101:taylor_ext}
    \begin{aligned}
    &\sum_{i=1}^{\vert\mathcal{V}\vert}\mathcal{M}_{i}\left(\mathbf{h}^{(k)}_{i}-\eta\nabla\mathcal{M}_{i}\left(\mathbf{h}^{(k)}_{i}\right)\right)=\sum_{i=1}^{\vert\mathcal{V}\vert} \Big\{\mathcal{M}_{i}\left(\mathbf{h}^{(k)}_{i}\right)-\\ &~~~~~~~~\Big[\eta\nabla\mathcal{M}_{i}\left(\mathbf{h}^{(k)}_{i}\right)\Big]^{\top}\int_0^1\nabla\mathcal{M}_{i}\left(\mathbf{h}^{(k)}_{i}-\theta\eta\nabla\mathcal{M}_{i}\left(\mathbf{h}^{(k)}_{i}\right)\right)\mathrm{d}\theta\Big\}\\
    &~~~=\sum_{i=1}^{\vert\mathcal{V}\vert} \Big\{\mathcal{M}_{i}\left(\mathbf{h}^{(k)}_{i}\right)-\Big[\eta\nabla\mathcal{M}_{i}\left(\mathbf{h}^{(k)}_{i}\right)\Big]^{\top}\nabla\mathcal{M}_{i}\left(\mathbf{h}^{(k)}_{i}\right)-\Big[\eta\nabla\mathcal{M}_{i}\left(\mathbf{h}^{(k)}_{i}\right)\Big]^{\top}\\
    &~~~~~~~~\times\int_0^1\left[\nabla\mathcal{M}_{i}\left(\mathbf{h}^{(k)}_{i}-\theta\eta\nabla\mathcal{M}_{i}\left(\mathbf{h}^{(k)}_{i}\right)\right)-\nabla\mathcal{M}_{i}\left(\mathbf{h}^{(k)}_{i}\right)\right]\mathrm{d}\theta\Big\}\\
    &~~~\leq\sum_{i=1}^{\vert\mathcal{V}\vert} \Big\{\mathcal{M}_{i}\left(\mathbf{h}^{(k)}_{i}\right)-\Big[\eta\nabla\mathcal{M}_{i}\left(\mathbf{h}^{(k)}_{i}\right)\Big]^{\top}\nabla\mathcal{M}_{i}\left(\mathbf{h}^{(k)}_{i}\right)+\\
    &~~~~~~~~\int_0^1\left\|\eta\nabla\mathcal{M}_{i}\left(\mathbf{h}^{(k)}_{i}\right)\right\|\left\|\nabla\mathcal{M}_{i}\left(\mathbf{h}^{(k)}_{i}-\theta\eta\nabla\mathcal{M}_{i}\left(\mathbf{h}^{(k)}_{i}\right)\right)-\nabla\mathcal{M}_{i}\left(\mathbf{h}^{(k)}_{i}\right)\right\|\mathrm{d}\theta\Big\}.\\
    \end{aligned}
\end{equation}
Note that 
\begin{equation}
    \label{ssy1101:update_equals_nabla}
    \begin{aligned}
    \mathbf{h}^{(k)}_{i}-\frac{1}{2}\nabla\mathcal{M}_{i}\left(\mathbf{h}^{(k)}_{i}\right)&=\mathbf{h}^{(k)}_{i}-\sum_{j=1}^{N}\left[\frac{\mathbf{A}(i,j)}{\mathbf{D}(i,i)}\mathbf{h}^{(k)}_{i}-\frac{\mathbf{A}(i,j)}{\mathbf{D}(j,j)}\mathbf{h}^{(k)}_{j}\right]=\mathbf{h}^{(k+1)}_{i},
    \end{aligned}
\end{equation}
the Eq.~\eqref{ssy1101:taylor_ext} can be further expressed as:
\begin{equation}
    \label{ssy1101:taylor_ext_further}
    \begin{aligned}
    &\sum_{i=1}^{\vert\mathcal{V}\vert}\mathcal{M}_{i}\left(\mathbf{h}^{(k)}_{i}-\eta\nabla\mathcal{M}_{i}\left(\mathbf{h}^{(k)}_{i}\right)\right)\leq\sum_{i=1}^{\vert\mathcal{V}\vert} \Bigg\{\mathcal{M}_{i}\left(\mathbf{h}^{(k)}_{i}\right)-\Big[\eta\nabla\mathcal{M}_{i}\left(\mathbf{h}^{(k)}_{i}\right)\Big]^{\top}\nabla\mathcal{M}_{i}\left(\mathbf{h}^{(k)}_{i}\right)\\
    &~~~~~~~~+\int_0^1\left\|\eta\nabla\mathcal{M}_{i}\left(\mathbf{h}^{(k)}_{i}\right)\right\|\left\|2\theta\eta\left(1-\frac{\mathbf{A}(i,i)}{\mathbf{D}(i,i)}\right)\nabla\mathcal{M}_{i}\left(\mathbf{h}^{(k)}_{i}\right)\right\|\mathrm{d}\theta\Bigg\}\\
    &~~~\leq\sum_{i=1}^{\vert\mathcal{V}\vert}\Big[\mathcal{M}_{i}\left(\mathbf{h}^{(k)}_{i}\right)-\left(\eta-\eta^2\right)\left\|\nabla\mathcal{M}_{i}\left(\mathbf{h}^{(k)}_{i}\right)\right\|^2\Big].
    \end{aligned}
\end{equation}
\indent The Eq. \eqref{ssy1101:taylor_ext_further} indicates that when the value of $\eta$ changes between 0 and 1, the inequality $\sum_{i=1}^{\vert\mathcal{V}\vert}\mathcal{M}_{i}\left(\mathbf{h}^{(k)}_{i}-\eta\nabla\mathcal{M}_{i}\left(\mathbf{h}^{(k)}_{i}\right)\right)\leq$$\sum_{i=1}^{\vert\mathcal{V}\vert}\mathcal{M}_{i}\left(\mathbf{h}^{(k)}_{i}\right)$ always holds. We have shown in Eq.~\eqref{ssy1101:update_equals_nabla} that $\mathbf{h}^{(k+1)}_{i}=\mathbf{h}^{(k)}_{i}-\frac{1}{2}\nabla\mathcal{M}_{i}\left(\mathbf{h}^{(k)}_{i}\right)$, so Eq. \eqref{ssy1101:taylor_ext_further} at $\eta=\frac{1}{2}$ can be written as:
\begin{equation}
\begin{aligned}
\sum_{i=1}^{\vert\mathcal{V}\vert}\mathcal{M}_{i}\left(\mathbf{h}^{(k+1)}_{i}\right)&=\sum_{i=1}^{\vert\mathcal{V}\vert}\mathcal{M}_{i}\left(\mathbf{h}^{(k)}_{i}-\frac{1}{2}\nabla\mathcal{M}_{i}\left(\mathbf{h}^{(k)}_{i}\right)\right)\leq\sum_{i=1}^{\vert\mathcal{V}\vert}\mathcal{M}_{i}\left(\mathbf{h}^{(k)}_{i}\right). 
\end{aligned}
\end{equation} 
This implies that the variation of node embeddings decreases (shrinking) as the message passing proceeds. In other words, embeddings of nodes shrink into distinct clusters in the space based on the graph topology. The motivation of the proposed methods is to eliminate unreliable connections in the graph topology, allowing related nodes (e.g., nodes of the same class) to be connected as much as possible. This enables the embeddings to shrink into distinct clusters according to their classes, thus clarifying the classification boundaries and enhancing the model's performance (we will verify this in subsequent experiments). 

\section{Experiments}\label{sec:experiment}
\subsection{Datasets and baselines} Experiments are conducted on four real-world TAGs: Cora~\cite{DBLP:journals/IR/McCallumNigam20}, Citeseer~\cite{DBLP:conf/DL/GilesBollacker98}, Pubmed~\cite{DBLP:journals/AIMagz/SenNamata08}, Arxiv-2023~\cite{DBLP:conf/iclr/HeBresson24}. 
We evaluate our methods by comparing them against 13 baselines from the following three categories: (1) Classical GNNs: GCN~\cite{DBLP:conf/iclr/Thomas16}, GAT~\cite{DBLP:conf/iclr/VelickovicCucurull18}, GraphSAGE~\cite{DBLP:conf/nips/HamiltonYing17}; (2) LM-based methods: BERT~\cite{DBLP:conf/NAACL/DevlinChang19}, DeBERTa~\cite{arxiv:HeLiu20}, GIANT~\cite{DBLP:conf/iclr/ChienChang22}, and variants combining classical GNN with DeBERTa encoding (GCN$_{\rm DeBERTa}$, GAT$_{\rm DeBERTa}$, GCN-LPA$_{\rm DeBERTa}$); (3) LLM-based methods: GPT-3.5-Turbo~\cite{DBLP:conf/nips/TomB20}, TAPE~\cite{DBLP:conf/iclr/HeBresson24} (the SOTA paradigm combining LLMs and GNNs, including TAPE$_{\rm GAT}$, TAPE$_{\rm GCN}$, and TAPE$_{\rm GCN-LPA}$).


\subsection{Implementation Details}
We adopt the GCN as the underlying graph learning architecture to validate the effect of LLM-based topology perturbations. The hyperparameters used for the GCN are consistent with previous studies~\cite{DBLP:conf/iclr/HeBresson24,arxiv:HuFey20}. As our goal is to investigate the effects of topology modifications, the hyperparameters associated with the GCN structure are not tuned specifically for each dataset. Considering the expenses associated with querying LLMs' APIs and the rate limit imposed by OpenAI, we randomly select 1,000 edges as candidate edges for edge deletion/addition in each dataset. For edge deletion, we choose candidate edges from the existing edges in each dataset. For edge addition, we select candidate edges based on the second-order neighbors of the nodes. Specifically, if two nodes are second-order neighbors and there is no existing edge between them, these two nodes will be considered as candidate node pairs for edge addition.
The threshold $\xi$ used in the deletion/addition process is tuned between 0.1 and 0.9 with an interval of 0.1. The threshold $\xi$ is tuned between 0.1 and 0.9 (interval 0.1). The $\lambda$ and $\beta$ are tuned between 0 and 5 (interval 0.1). For the general setting, 60\% of the nodes are designated for the training set, 20\% for the validation set, and the remaining 20\% are set aside for the test set. For the few-shot setting, we randomly select 20 nodes from each class to form the training set, 500 random nodes for the validation set, and 1,000 random nodes from the remaining pool for the test set. 

\begin{table*}[t]
  \centering
  \caption{Experimental results in the general setting (\textbf{Best}, \underline{second Best}).}
  \scalebox{0.75}{
    \begin{tabular}{ccccc}
    \toprule[1.5pt]
    Dataset & Cora  & Pubmed & Citeseer & Arxiv-2023 \\
    \midrule
    GPT-3.5-Turbo   & 0.6769 ± 0.0000 & 0.9342 ± 0.0000 & 0.5929 ± 0.0000 & 0.7356 ± 0.0000 \\
    GAT   &  0.8573 ± 0.0065 & 0.8328 ± 0.0012  &  0.7423 ± 0.0178 & 0.6784 ± 0.0023 \\
    GCN   & 0.8690 ± 0.0151 & 0.8890 ± 0.0032 & 0.7298 ± 0.0132  & 0.6760 ± 0.0028 \\
    GraphSAGE   & 0.8573 ± 0.0065 & 0.8685 ± 0.0011 & 0.7361 ± 0.0190  &  0.6906 ± 0.0024\\
    GIANT & 0.8423 ± 0.0053 & 0.8419 ± 0.0050 & 0.7238 ± 0.0083 & 0.5672 ± 0.0061 \\
    BERT  & 0.7400 ± 0.0175 & 0.9058 ± 0.0046 & 0.7317 ± 0.0175 & 0.6840 ± 0.0122 \\
    DeBERTa   &  0.7385 ± 0.0127 & 0.9020 ± 0.0057 & 0.7313 ± 0.0194  &  0.6789 ± 0.0185 \\
    GAT$_{\rm DeBERTa}$   & 0.8750 ± 0.0084 & 0.9312 ± 0.0083 & 0.7547 ± 0.0231 & 0.7704 ± 0.0043 \\
    GCN$_{\rm DeBERTa}$   & 0.8778 ± 0.0137 & 0.9314 ± 0.0039 & 0.7508 ± 0.0066  & 0.7694 ± 0.0022 \\
    GCN-LPA$_{\rm DeBERTa}$ & 0.8750 ± 0.0209 & 0.9446 ± 0.0030 & 0.7559 ± 0.0104  & 0.7831 ± 0.0038 \\
    GCN + LLM-based A-D &  \cellcolor{gray!15}\underline{0.8815 ± 0.0180} & \cellcolor{gray!15}0.9309 ± 0.0050 & \cellcolor{gray!15}0.7578 ± 0.0079 & \cellcolor{gray!15}0.7708 ± 0.0009 \\
    GCN + LLM-based LPA & \cellcolor{gray!15}\textbf{ 0.8828 ± 0.0191} &  \cellcolor{gray!15}\underline{0.9469 ± 0.0037} &  \cellcolor{gray!15}\underline{0.7614 ± 0.0149} &  \cellcolor{gray!15}\underline{0.7853 ± 0.0022} \\
    GCN + LLM-based A-D \& LPA & \cellcolor{gray!15} 0.8778 ± 0.0183 & \cellcolor{gray!15}\textbf{0.9475 ± 0.0036} & \cellcolor{gray!15}\textbf{0.7633 ± 0.0117} & \cellcolor{gray!15}\textbf{0.7858 ± 0.0027}  \\
    \midrule
    $\rm{TAPE_{GAT}}$   & 0.8838 ± 0.0088 & 0.9316 ± 0.0115 & 0.7610 ± 0.0107 & 0.7807 ± 0.0023 \\
    $\rm{TAPE_{GCN}}$   & 0.8833 ± 0.0046 & 0.9319 ± 0.0037 & 0.7571 ± 0.0046 & 0.7819 ± 0.0027 \\
    $\rm{TAPE_{GCN-LPA}}$ & 0.8824 ± 0.0051 & 0.9461 ± 0.0023 & 0.7653 ± 0.0073 & 0.8020 ± 0.0029 \\
    $\rm{TAPE_{GCN}}$ + LLM-based A-D &  \cellcolor{gray!15}\underline{0.8907 ± 0.0138} & \cellcolor{gray!15}0.9329 ± 0.0045 & \cellcolor{gray!15}0.7625 ± 0.0101 & \cellcolor{gray!15}0.7840 ± 0.0012 \\
    $\rm{TAPE_{GCN}}$ + LLM-based LPA &  \cellcolor{gray!15}0.8875 ± 0.0151 & \cellcolor{gray!15}\textbf{0.9475 ± 0.0049} & \cellcolor{gray!15}\textbf{0.7692 ± 0.0059} & \cellcolor{gray!15}\textbf{0.8033 ± 0.0037} \\
    $\rm{TAPE_{GCN}}$ + LLM-based A-D \& LPA & \cellcolor{gray!15}\textbf{0.8916 ± 0.0096} &  \cellcolor{gray!15}\underline{0.9472 ± 0.0056} &  \cellcolor{gray!15}\underline{0.7684 ± 0.0076} &  \cellcolor{gray!15}\underline{0.8032 ± 0.0023} \\
    \bottomrule[1.5pt]
    \end{tabular}}
  \label{ssy1101:general}%
\end{table*}%

\begin{table*}[t]
  \centering
  \caption{Experimental results in the few-shot setting (\textbf{Best}, \underline{second Best}).}
  \scalebox{0.8}{
    \begin{tabular}{ccccc}
    \toprule[1.5pt]
    Dataset & Cora  & Pubmed & Citeseer & Arxiv-2023 \\
    \midrule
    GAT$_{\rm DeBERTa}$   & 0.7272 ± 0.0315 & 0.6462 ± 0.0685  & \underline{0.6703 ± 0.0429} & 0.4185 ± 0.1070 \\
    GCN$_{\rm DeBERTa}$   & 0.7552 ± 0.0094 & \underline{0.6462 ± 0.0664} & 0.6647 ± 0.0337 & 0.4960 ± 0.0458 \\
    GCN-LPA$_{\rm DeBERTa}$ & \underline{0.7588 ± 0.0132} & 0.6382 ± 0.0550 & 0.6657 ± 0.0319 & \underline{0.5050 ± 0.0437} \\
    GCN + LLM-based LPA & \cellcolor{gray!15}{\textbf{0.7630 ± 0.0134}} & \cellcolor{gray!15}{\textbf{0.6535 ± 0.0427}} & \cellcolor{gray!15}{\textbf{0.6780 ± 0.0341}} & \cellcolor{gray!15}{\textbf{0.5175 ± 0.0392}} \\
    \midrule
    $\rm{TAPE_{GAT}}$   & 0.7990 ± 0.0054 & 0.8573 ± 0.0107 & \underline{0.7223 ± 0.0240} & 0.6747 ± 0.0826 \\
    $\rm{TAPE_{GCN}}$   & 0.8190 ± 0.0050 & 0.8668 ± 0.0147  & 0.7200 ± 0.0168 & 0.7365 ± 0.0271 \\
    $\rm{TAPE_{GCN-LPA}}$ & \underline{0.8232 ± 0.0033} & \underline{0.8815 ± 0.0195} & 0.7198 ± 0.0084 & \underline{0.7382 ± 0.0278} \\
    $\rm{TAPE_{GCN}}$ + LLM-based LPA & \cellcolor{gray!15}{\textbf{0.8275 ± 0.0066}} & \cellcolor{gray!15}{\textbf{0.8930 ± 0.0256}} & \cellcolor{gray!15}{\textbf{0.7307 ± 0.0143}} & \cellcolor{gray!15}{\textbf{0.7428 ± 0.0272}} \\
    \bottomrule[1.5pt]
    \end{tabular}}
  \label{ssy1101:few_shot}%
\end{table*}%

\begin{figure}
    \centering
    \begin{subfigure}[b]{0.493\textwidth}
        \centering
        \includegraphics[width=\linewidth]{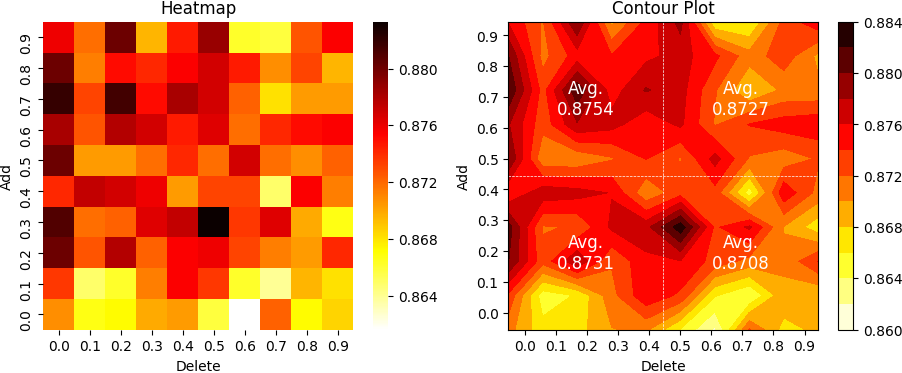}
        \caption{Cora dataset.}
        \label{fig:subfig_a}
    \end{subfigure}
    \begin{subfigure}[b]{0.493\textwidth}
        \centering
        \includegraphics[width=\linewidth]{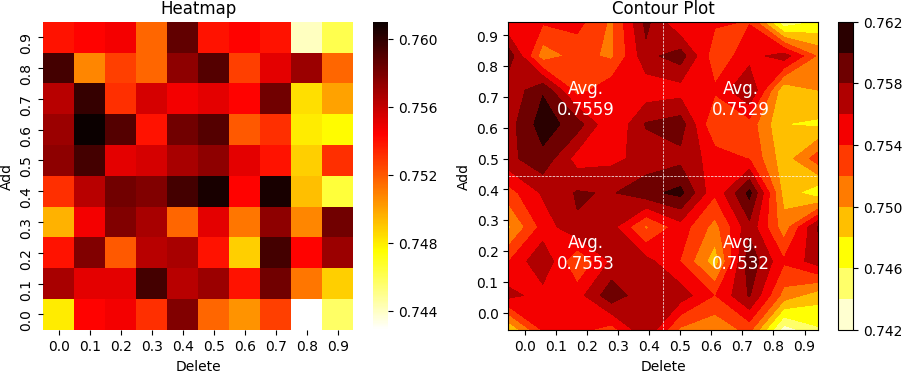}
        \caption{Citeseer dataset.}
        \label{fig:subfig_b}
    \end{subfigure}
    \caption{Heatmaps and contour plots on different datasets.}
    \label{ssy1101:heatmap_and_contour}
\end{figure}

\begin{figure}[h]
    \centering
    \begin{subfigure}[b]{0.23\textwidth}
        \centering
        \includegraphics[width=\linewidth]{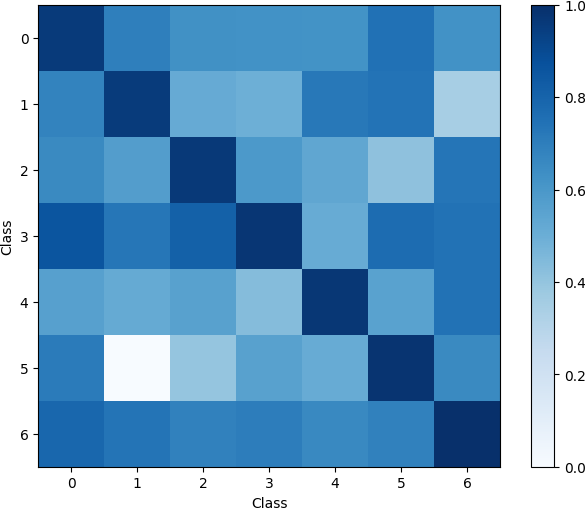}
        \caption{Cora.}
        \label{fig:subfig_a}
    \end{subfigure}
    \begin{subfigure}[b]{0.23\textwidth}
        \centering
        \includegraphics[width=\linewidth]{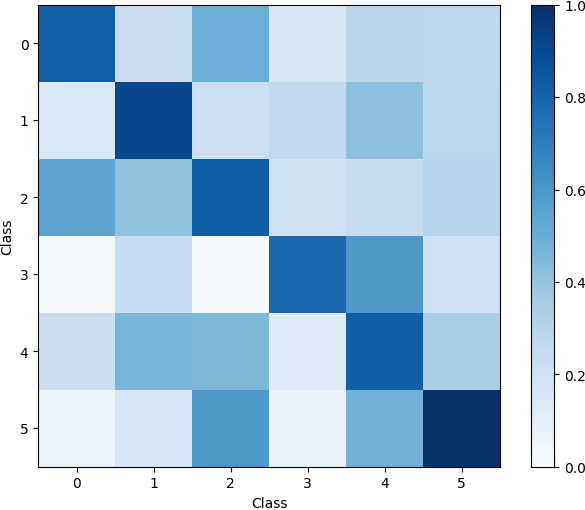}
        \caption{Citeseer.}
        \label{fig:subfig_b}
    \end{subfigure}
    \begin{subfigure}[b]{0.23\textwidth}
        \centering
        \includegraphics[width=\linewidth]{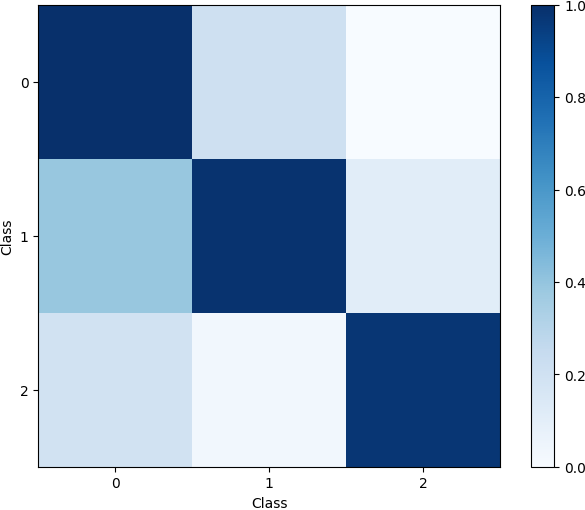}
        \caption{Pubmed.}
        \label{fig:subfig_b}
    \end{subfigure}
    \begin{subfigure}[b]{0.23\textwidth}
        \centering
        \includegraphics[width=\linewidth]{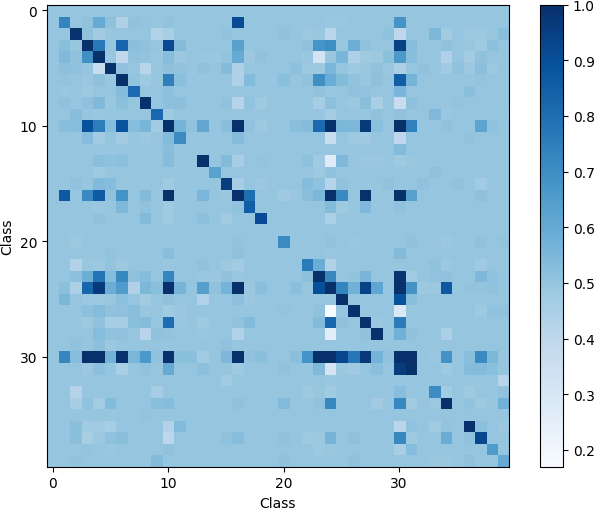}
        \caption{Arxiv-2023.}
        \label{fig:subfig_b}
    \end{subfigure}
    \caption{The relationship between the refined edge weights and classes.}
    \label{ssy1101:learned_egde_weight}
\end{figure}

\begin{figure}[h]
    \centering
    \begin{subfigure}[b]{0.26\textwidth}
        \centering
        \includegraphics[width=\linewidth]{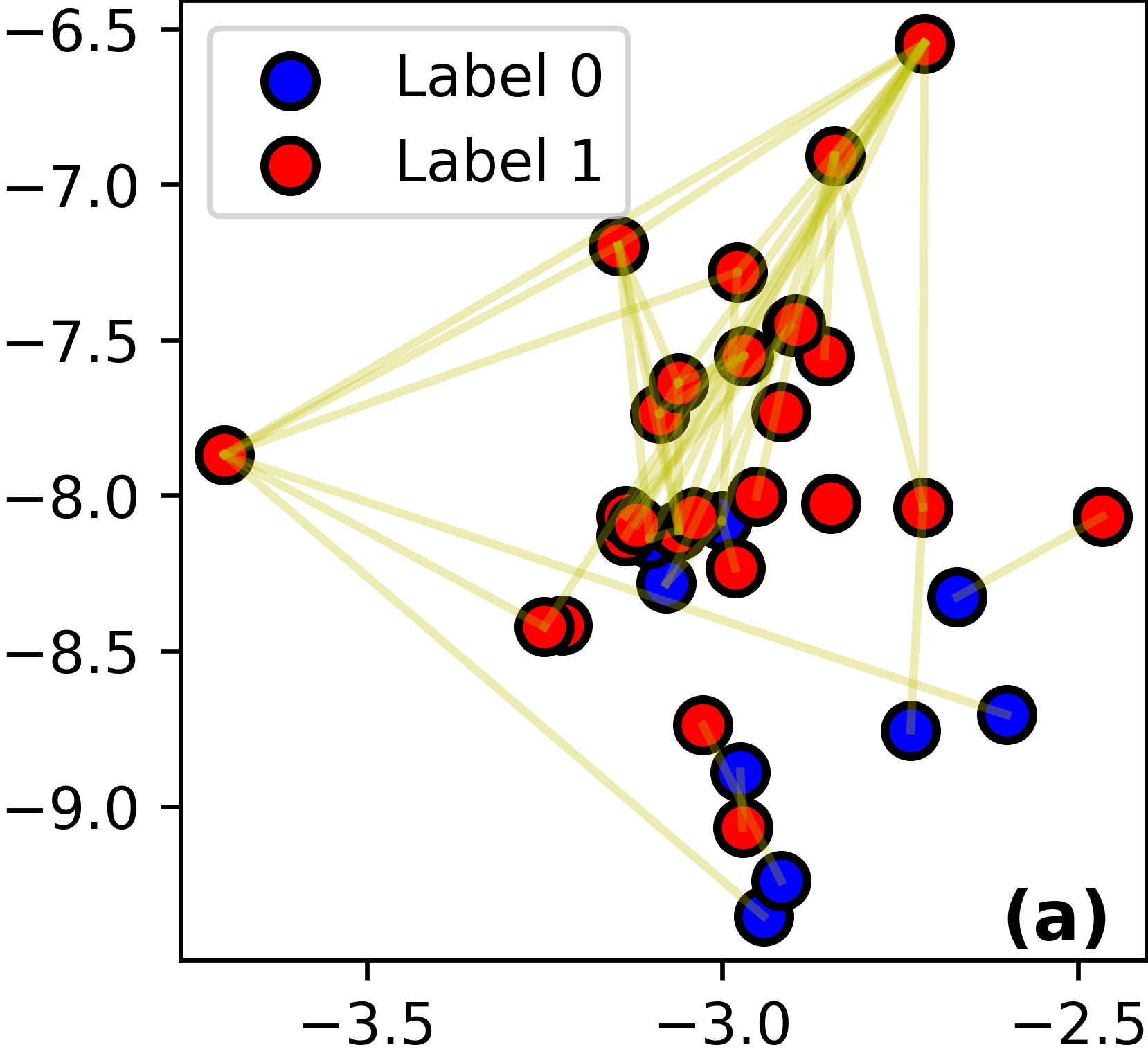}
        \label{fig:subfig_a}
    \end{subfigure}\hspace{1em}
    \begin{subfigure}[b]{0.26\textwidth}
        \centering
        \includegraphics[width=\linewidth]{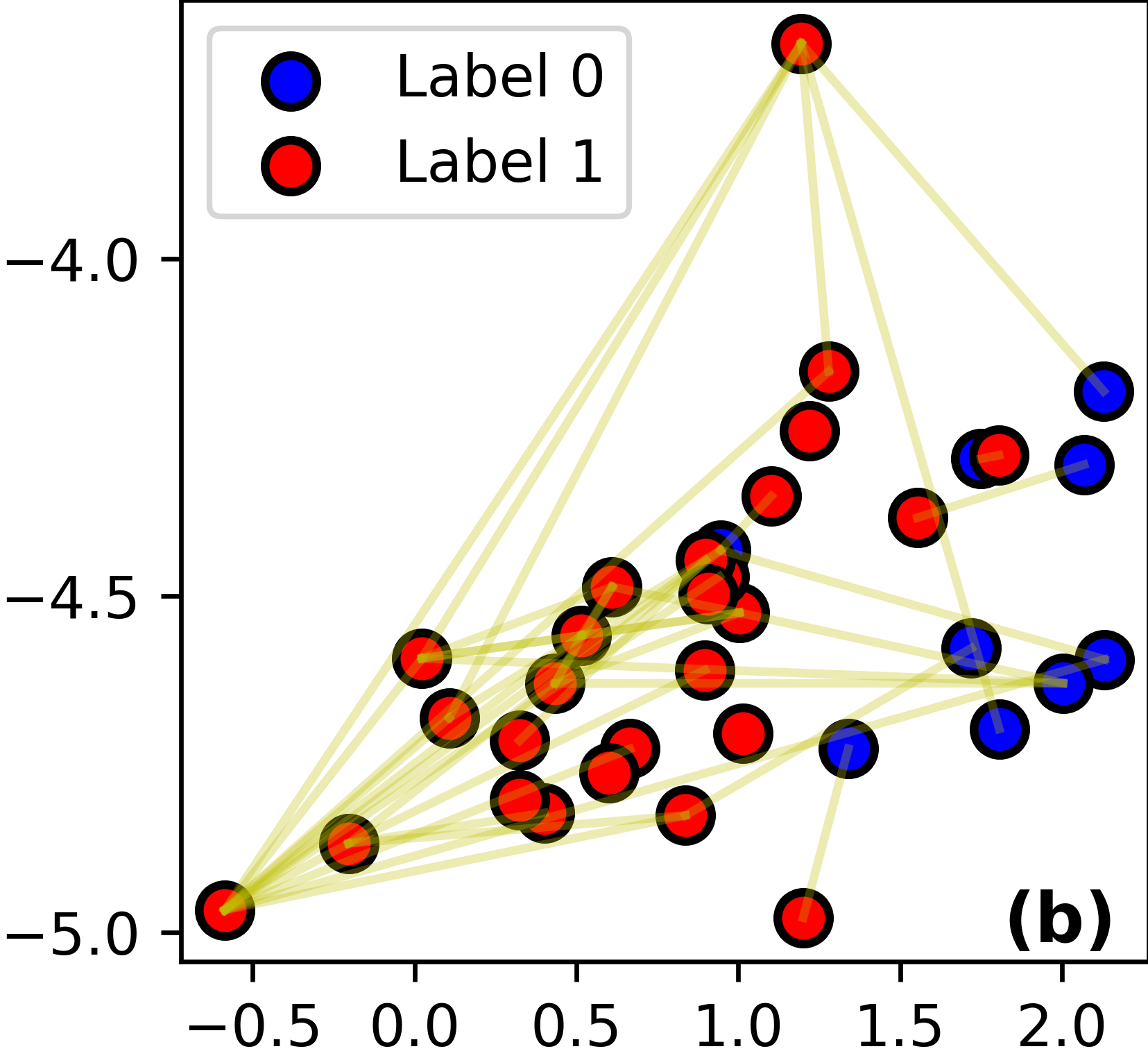}
        \label{fig:subfig_b}
    \end{subfigure}\hspace{1em}
    \begin{subfigure}[b]{0.26\textwidth}
        \centering
        \includegraphics[width=\linewidth]{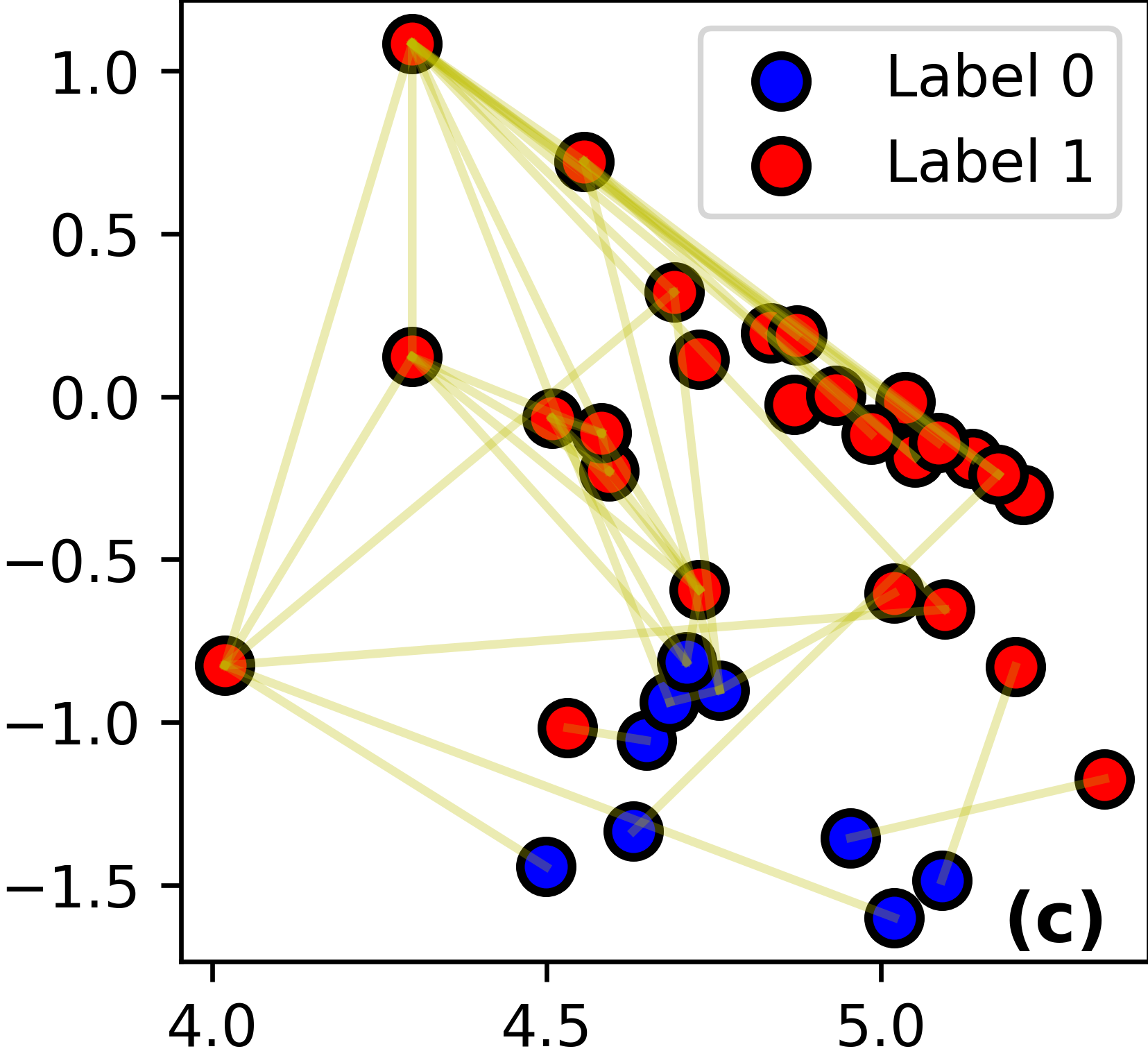}
        \label{fig:subfig_b}
    \end{subfigure}
    \caption{Discriminativeness of learned embeddings. (a) $w/o$ topology enhancement. (b) $w/$ LLM-based A-D. (c) $w/$ LLM-based LPA.}
    \label{ssy1101:learned_embedding}
\end{figure}

\begin{figure}[h]
    \centering
    \begin{subfigure}[b]{0.43\textwidth}
        \centering
        \includegraphics[width=\linewidth]{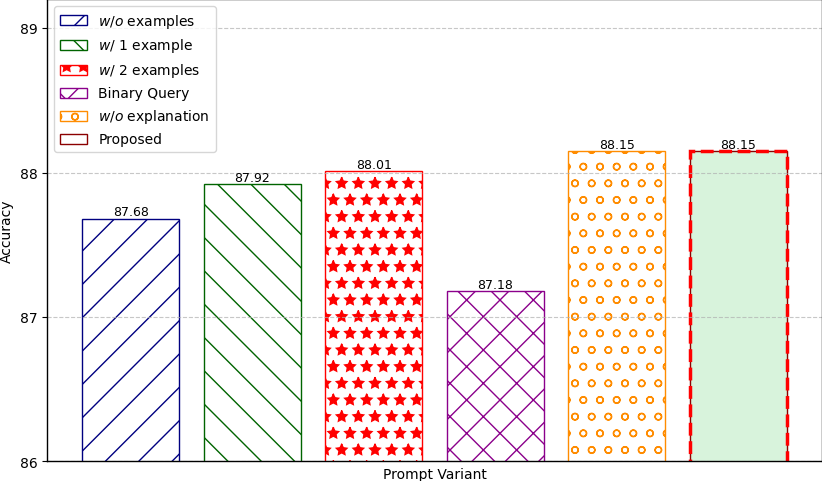}
        \caption{}
        \label{fig:subfig_a}
    \end{subfigure}
    \begin{subfigure}[b]{0.43\textwidth}
        \centering
        \includegraphics[width=\linewidth]{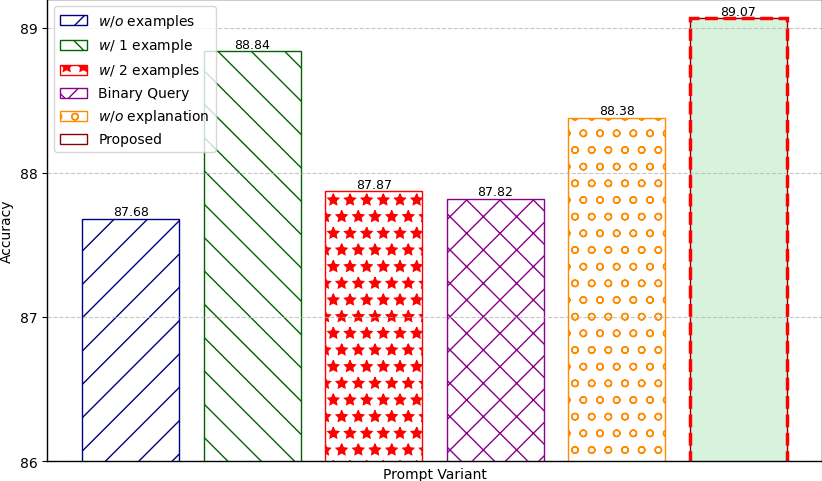}
        \caption{}
        \label{fig:subfig_b}
    \end{subfigure}
    \caption{Prompt variant analysis. (a) Traditional paradigm. (b) TAPE paradigm.}
    \label{ssy1101:prompt_variant}
\end{figure}

\subsection{Performance Comparisons}\label{ssy1101:perform_comp}
\textbf{General setting}. Table~\ref{ssy1101:general} shows the experimental results under the general setting, where ``LLM-based A-D'' represents ``LLM-based Edge Deletion and Edge Addition'', ``LLM-based LPA'' represents ``LLM-based Pseudo-label Propagation'', and ``LLM-based A-D \& LPA'' represents the combination of ``LLM-based A-D'' and ``LLM-based LPA''. It can be seen that the LLM-based topological structure perturbation has certain positive effects. The ``GCN + LLM-based A-D'' can achieve performance gains compared to the GCN (or GCN$_{\rm DeBERTa}$) in most cases. This is attributed to the powerful text understanding ability of the LLM, which enables it to filter out unreliable edges in the graph and add reliable ones
, thereby enhancing the quality of the learned representations (through facilitating the process of message passing). Similar performance gains can be observed in the ``GCN + LLM-based LPA''. Compared to the classical label propagation method like GCN-LPA$_{\rm DeBERTa}$, the LLM-based pseudo-label propagation introduces more label information by utilizing the powerful reasoning capabilities of the LLM. Specifically, the GCN-LPA$_{\rm DeBERTa}$ only uses the labels of labeled nodes (unlabeled nodes use a default label) during label propagation, whereas the ``GCN + LLM-based LPA'' not only uses the labels of labeled nodes, but also high-quality pseudo-labels generated by the LLM for unlabeled nodes. The additional label information aids label propagation, which in turn helps guide the model to learn more proper edge weights (i.e., a better graph topology). Although the LLM-based edge deletion/addition and the LLM-based pseudo-label propagation have differences in the direction of optimizing the graph topology, we find that combining the two can further enhance the model performance. As shown in the Table~\ref{ssy1101:general}, the ``GCN + LLM-based A-D \& LPA'' is almost always the top performer across all settings under the traditional paradigm. We also verify the generalizability of the refined graph topology structure on the SOTA paradigm TAPE, i.e., using the ``TA+P+E'' feature~\cite{DBLP:conf/iclr/HeBresson24}. ``TA+P+E'' feature is a recently proposed LLM-augmented feature that incorporates the original features of the TAG (TA), LLM explanations (E), and LLM predictions (P). The results demonstrate that the LLM-based graph topology refinement approaches also benefit learning with the ``TA+P+E'' feature, showcasing the versatility of the proposed methods to some extent.

\spara{Few-shot setting} To further demonstrate the impact of pseudo-labels generated by the LLM on the label propagation, we test the performance of the ``GCN + LLM-based LPA'' in the few-shot setting, as shown in the Table~\ref{ssy1101:few_shot}. The ``GCN + LLM-based LPA'' has achieved performance gains in different scenarios, which is consistent with the experimental results under the general setting. We also test the generalizability of the optimized graph topology structure on the SOTA paradigm TAPE. The experimental results demonstrate that the LLM-based graph topology perturbation also improves learning with the ``TA+P+E'' feature under the few-shot setting.

\subsection{Visualization and Analysis}
\spara{Visualization of LLM's topology refinement capability} In Fig.~\ref{ssy1101:heatmap_and_contour}, we present heatmaps and contour plots on different datasets, which are obtained by adjusting the thresholds used in the edge deletion and addition processes. From these heatmaps and contour plots, we can observe a consistent pattern: the values in the top left corner of the figures are relatively large, while the values in the bottom right corner are relatively small. In other words, adding edges that LLM considers reliable and deleting those are considered as unreliable is beneficial to improving the topological structure, thereby improving the classification accuracy. For example, in the contour plot of the Cora, the average accuracy in the top left corner is 0.8754, while in the bottom right corner it is 0.8708. This phenomenon indicates that LLMs do have the ability to make reasonable edge additions/deletions based on the semantics of the input text.

\spara{Analysis of learned edge weights and node embeddings} In Fig.~\ref{ssy1101:learned_egde_weight}, we visualize the relationship between the refined edge weights and classes. Specifically, we divide the nodes into different groups according to their classes and then calculate the average edge weights between different groups. It is evident from Fig.~\ref{ssy1101:learned_egde_weight} that the average weights on the diagonal are the highest. This indicates that the proposed model tends to strengthen the weights between nodes of the same class (i.e., emphasizing reliable edges in message passing) to achieve the purpose of topology enhancement. We also show the learned embeddings in Fig.~\ref{ssy1101:learned_embedding}. We sample nodes of two different classes from the Cora, and project the embeddings of these nodes onto a 2D plane using t-SNE. It can be observed that the embeddings with topology enhancement exhibit better discrimination, which aligns with the theoretical analysis (shrinking theory) provided in Section~\ref{ssy1101:theory_analysis}.

\begin{figure}[t]
    \centering
    \begin{subfigure}[b]{0.45\textwidth}
        \centering
        \includegraphics[width=\linewidth]{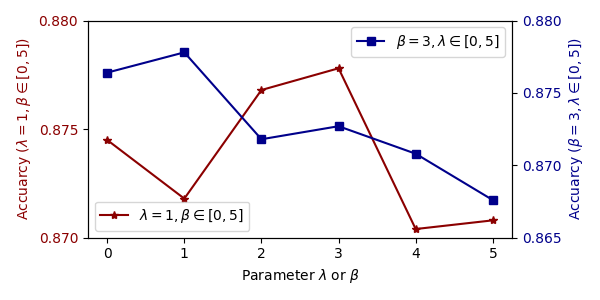}
        \caption{General setting.}
        \label{fig:subfig_a}
    \end{subfigure}
    \begin{subfigure}[b]{0.45\textwidth}
        \centering
        \includegraphics[width=\linewidth]{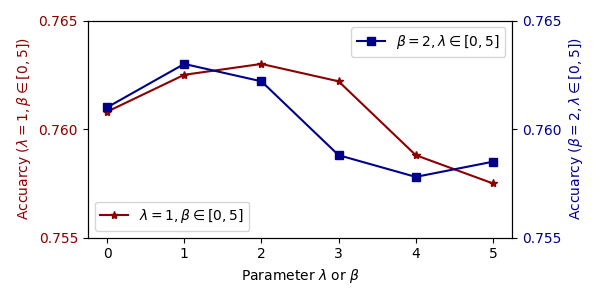}
        \caption{Few-shot setting.}
        \label{fig:subfig_b}
    \end{subfigure}
    \caption{Parameter sensitivity of $\lambda$ and $\beta$.}
    \label{ssy1101:param_sensitivity}
\end{figure}

\spara{Sensitivity evaluation of prompts and parameters} In Fig.~\ref{ssy1101:prompt_variant}, we analyze different variants of prompts, including using fewer examples ($w/o$ examples, $w/$ 1 example, and $w/$ 2 examples), directly asking the LLM to determine whether an edge should exist (YES or NO, binary query), and not requiring the LLM to provide an explanation for its decisions ($w/o$ explanation). The experimental results indicate that using fewer examples degrades performance, for more examples help the LLM acquire task-specific knowledge. Additionally, binary queries and explanation-free modes are also detrimental to performance. This is because the fine-grained threshold queries and decision attributions in the proposed prompt template (Section \ref{sun1101:Methodology}) increase the reliability of the LLM's responses. Without them, the likelihood of LLM hallucinations will increase. The parameter sensitivity analysis is shown in Fig.~\ref{ssy1101:param_sensitivity}. The phenomenon shown in Fig.~\ref{ssy1101:param_sensitivity} indicates that the modules corresponding to both hyperparameters have positive effects on the final model performance. The absence of either one ($\lambda=0$ or $\beta=0$) results in a decrease in model performance.\\

\section{Conclusion}\label{sec:conclusion}
In this paper, we investigate the potential of LLMs in refining graph topology. First, we explore ways to leverage LLMs' capabilities in deleting unreliable edges and adding reliable edges in TAGs through the prompt statement. Second, considering LLMs' text understanding ability can help generate high-quality pseudo-labels, we propose to leverage the pseudo-labels produced by LLMs to improve the topological structure of TAGs. Finally, we integrate the above methods into the GNN training. The experimental results demonstrate that LLMs can play a positive role in perturbing the topological structure of graphs. 


\subsubsection{\ackname} This work was supported by the NSFC Young Scientists Fund (No. 9240127), the Donations for Research Projects (No. 9229129) of the City University of Hong Kong, the Early Career Scheme (No. CityU 21219323), and the General Research Fund (No. CityU 11220324) of the University Grants Committee (UGC).
\bibliographystyle{splncs04}
\bibliography{reference}

\end{document}